\documentclass[fleqn,10pt]{wlscirep}
\usepackage[utf8]{inputenc}
\usepackage[T1]{fontenc}
\usepackage{xcolor}
\usepackage{ulem}

\usepackage{authblk}
\usepackage{wrapfig}
\usepackage{ulem}
\usepackage{amssymb, amsfonts, amsmath, amsthm}
\usepackage{dsfont}
\usepackage{graphicx}
\usepackage{csvsimple}
\usepackage{breakcites}
\usepackage[ruled]{algorithm2e}
\usepackage{subfloat}
\usepackage{longtable}
\usepackage{float}
\usepackage{wrapfig}
\usepackage[section]{placeins}

\newcommand{\sref}[1]{Sec.~\ref{#1}}

\usepackage{amsmath, amsthm, amssymb, dsfont}
\title{Signed iterative random forests to identify enhancer-associated transcription factor binding}

\author[1]{Karl Kumbier}
\author[2]{Sumanta Basu}
\author[3]{Erwin Frise}
\author[3]{Susan E. Celniker}
\author[3,4,5]{James B. Brown}
\author[5,6,7*]{Bin Yu}

\affil[1]{University of California, San Francisco, Department of Pharmaceutical Chemistry, San Francisco, 94158, United States}
\affil[2]{Cornell University, Department of Statistics and Data Science, Ithaca, 14853, United States}
\affil[3]{Lawrence Berkeley National Biological Systems and Engineering Division, Berkeley, 94720, United States}
\affil[4]{Lawrence Berkeley National Laboratory, Division of Environmental Genomics and Systems Biology, Berkeley, 94720, United States}
\affil[5]{University of California, Berkeley, Department of Statistics, Berkeley, 94720, United States}
\affil[6]{University of California, Berkeley, Department of Electrical Engineering and Computer Science, Berkeley, 94720, United States}
\affil[7]{University of California, Berkeley, Center for Computational Biology, Berkeley, 94720, United States}

\affil[*]{binyu@berkeley.edu}

\begin{abstract}
Standard ChIP-seq peak calling pipelines seek to differentiate biochemically reproducible signals of individual genomic elements from background noise. However, reproducibility alone does not imply functional regulation (e.g., enhancer activation, alternative splicing). Here we present a general-purpose, interpretable machine learning method: signed iterative random forests (siRF), which we use to infer regulatory interactions among transcription factors and functional binding signatures surrounding enhancer elements in \textit{Drosophila melanogaster}.
\end{abstract}

\begin{document}

\flushbottom
\maketitle

\thispagestyle{empty}

Spatiotemporal gene expression patterns that coordinate embryonic development are regulated by combinatorial transcription factor (TF) interactions at enhancer elements\cite{hammonds2013spatial,gerstein2012architecture,basu2018iterative,arbel2019exploiting,roy2010identification,hoffman2012integrative}. Despite their importance, such interactions have only been well-characterized for a small number of enhancers. For example, the \textit{even-skipped (eve)} stripe 2 enhancer in \textit{Drosophila} drives expression in a narrow spatiotemporal window—a band three nuclei wide that forms and dissipates in less than an hour\cite{bothma2014dynamic}. This rapid process is governed by interactions among six TFs encoded by \textit{bicoid (bcd)}, \textit{giant (gt)}, \textit{hunchback (hb)}, \textit{Kr\"{u}ppel (Kr)} , \textit{stat92E}, and \textit{zelda}. The coordinated activity of these TFs defines and constrains the domain of \textit{eve} expression through thermodynamically coupled cooperative and competitive binding\cite{kim2013rearrangements}. 

Increasing availability of chromatin immunoprecipitation (ChIP) datasets\cite{macarthur2009developmental,encode2012integrated, celniker2009unlocking, kudron2018modern, zheng2019cistrome,roadmap2015integrative,li2014establishment} and high-throughput genomic annotations\cite{kvon2014genome,fishilevich2017genehancer,lesurf2016oreganno}  provide the means for data-driven, genome-wide discovery of TF interactions at enhancer elements. This challenge depends on two interrelated tasks: the binding of individual TFs must (i) be mapped across the genome and (ii) integrated into groups of TFs whose collective binding is associated with enhancer activation. Peak-calling algorithms\cite{kharchenko2008design, zhang2008model, landt2012chip} have previously been used, and paired with measures of biochemical reproducibility\cite{li2011measuring}, to accomplish (i). These pipelines have provided maps of biochemically reproducible,  genome-wide binding of individual TFs from ChIP-seq and ChIP-chip assays\cite{encode2012integrated, celniker2009unlocking, kudron2018modern}. While reproducibility is a useful criterion in general, it does not imply functional regulation \cite{li2008transcription,fisher2012dna,landt2012chip}.  For instance, in the MODERN Consortium, over 80\% of biochemically reproducible peaks were discarded based on their localization in “hot spots”, wherein many unrelated factors appear to bind—a putative indicator of artifacts\cite{kudron2018modern}. While these hot spots are highly reproducible, their functional status is debated\cite{macarthur2009developmental,fisher2012dna,kasinathan2014high,siersbaek2014molecular,moorman2006hotspots}. This debate is emblematic of the challenge with using the ChIP-seq assay (and related assays) to represent functional binding events. 

\begin{figure}
  \begin{center}
    \includegraphics[width=0.95\linewidth]{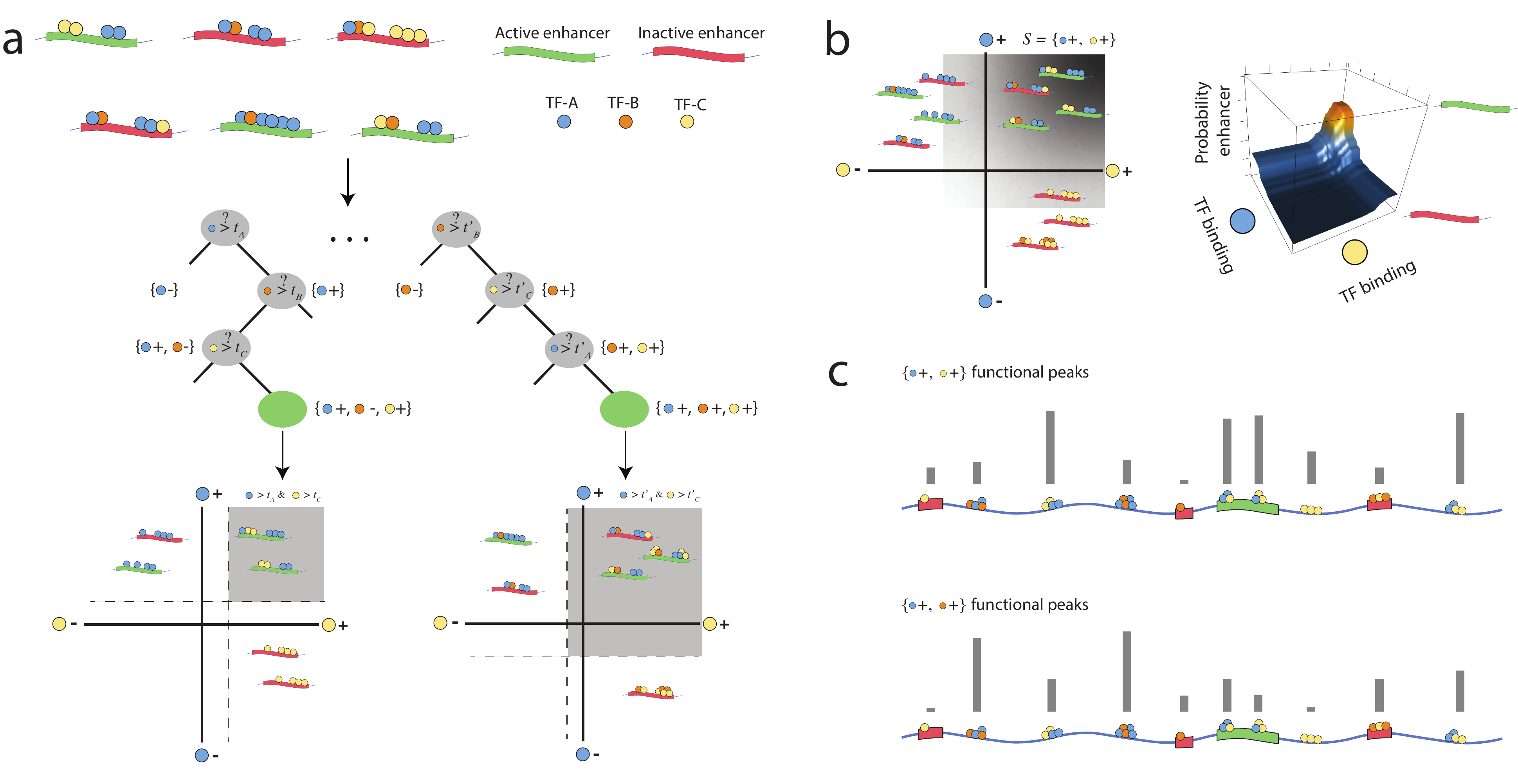}
  \end{center}
  \caption[siRF workflow]{Interaction discovery and functional peak prediction through siRF: \textbf{a.} Random forests learn predictive rules that differentiate between active and inactive enhancers. Each rule corresponds to a distinct pattern of TF enrichment/depletion (shown beside each decision tree node) and a region of the feature space that is enriched for active enhancers (shown below each decision tree node). \textbf{b.} Signed interactions $S$ relate similar rules throughout an RF and are associated with a subset of samples where these rules are active. For example, the signed interaction $S=\{Blue+, Yellow+\}$ relates rules defined by high levels of both the blue and yellow TFs. These rules are active for elements highlighted in grey region (left) and correspond to a model for predicting enhancer probability as a function of TF binding (right). \textbf{c}. TF binding models associated with signed interactions detected in \textbf{a-b} are used to generate genome-wide maps of putatively functional TF binding events. 
 } \label{fig:workflow}
\end{figure}

Here, we developed signed iterative random forests (siRF)—a general-purpose, interpretable machine learning algorithm (Fig. \ref{fig:workflow}, Methods)—that addresses (i) and (ii) in an integrated manner. SiRF refines the iterative random forest (iRF) algorithm\cite{basu2018iterative}, which identifies stable, high-order interactions in random forests (RF)\cite{breiman2001random}, to extract the underlying decision rules corresponding to these interactions. 
Training siRF to predict enhancer activity from raw ChIP signals allows us to systematically detect and rank patterns of TF enrichment/depletion, which we call \textit{signed interactions}, that are associated with enhancer activation. SiRF then estimates \textit{functional peaks} for each signed interaction—putative binding events that are statistically associated with a targeted functional element (e.g., enhancers)—using the decision rules associated with a given interaction. Thus, the functional peaks we identify provide hypotheses of both predicted enhancers and the regulatory interactions that drive enhancer activity. Intuitively, siRF re-frames peak calling as a problem of feature/interaction selection in supervised learning. In contrast, traditional methods (e.g.,  \cite{kharchenko2008design, zhang2008model, landt2012chip}) can be viewed as an unsupervised approach to peak calling. The peak sets from each approach may be used independently or in combination with one another, subject to accuracy (in the case of siRF) and reproducibility (in the case of traditional peak calling) considerations.

As a proof of principle, we used siRF to identify patterns of enriched/depleted TFs that are associated with enhancer activity during \textit{Drosophila} embryonic development. We analyzed a dataset of 7,705 genomic segments ($\sim$1-2kb), scored on a spectrum from non-enhancer to strong enhancer across 5 developmental stage windows\cite{kvon2014genome} (Supplementary Table 1), along with genome-wide ChIP-seq and ChIP-chip profiles for 262 unique TFs\cite{celniker2009unlocking,kudron2018modern,macarthur2009developmental} (Supplementary Table 2). A total of 15 siRF models were trained to predict enhancer activity (5 developmental stages $\times$ 3 enhancer scoring criteria; Methods). Prediction accuracy on held-out genomic segments was best for early-stage enhancers (stage 4-8 weak enhancer AUROC = 0.85, stage 4-8 strong enhancer AUROC = 0.94) and degraded at later stages (stage 15-16 weak enhancer AUROC = 0.68, stage 15-16 strong enhancer AUROC = 0.77; Supplementary Fig. S4-S5). From these 15 models, siRF identified 684 (505 unique over developmental stages) signed interactions (involving up to 4 TFs) that represent patterns of TF enrichment/depletion associated with enhancer activity (Supplementary Table 3). This corresponds to a $>99.9\%$ reduction in the $O(10^9)$ possible signed interactions (of up to order-4) among the 262 TFs included in our analysis. We note that the majority of TFs in our study have not been widely studied, and thus signed interactions serve as hypotheses for future investigation. However, most 457 / 684 of the interactions we discovered were from stage 4-8 models, where prediction accuracy was highest.

\begin{figure}
  \begin{center}
    \includegraphics[width=0.8\textwidth]{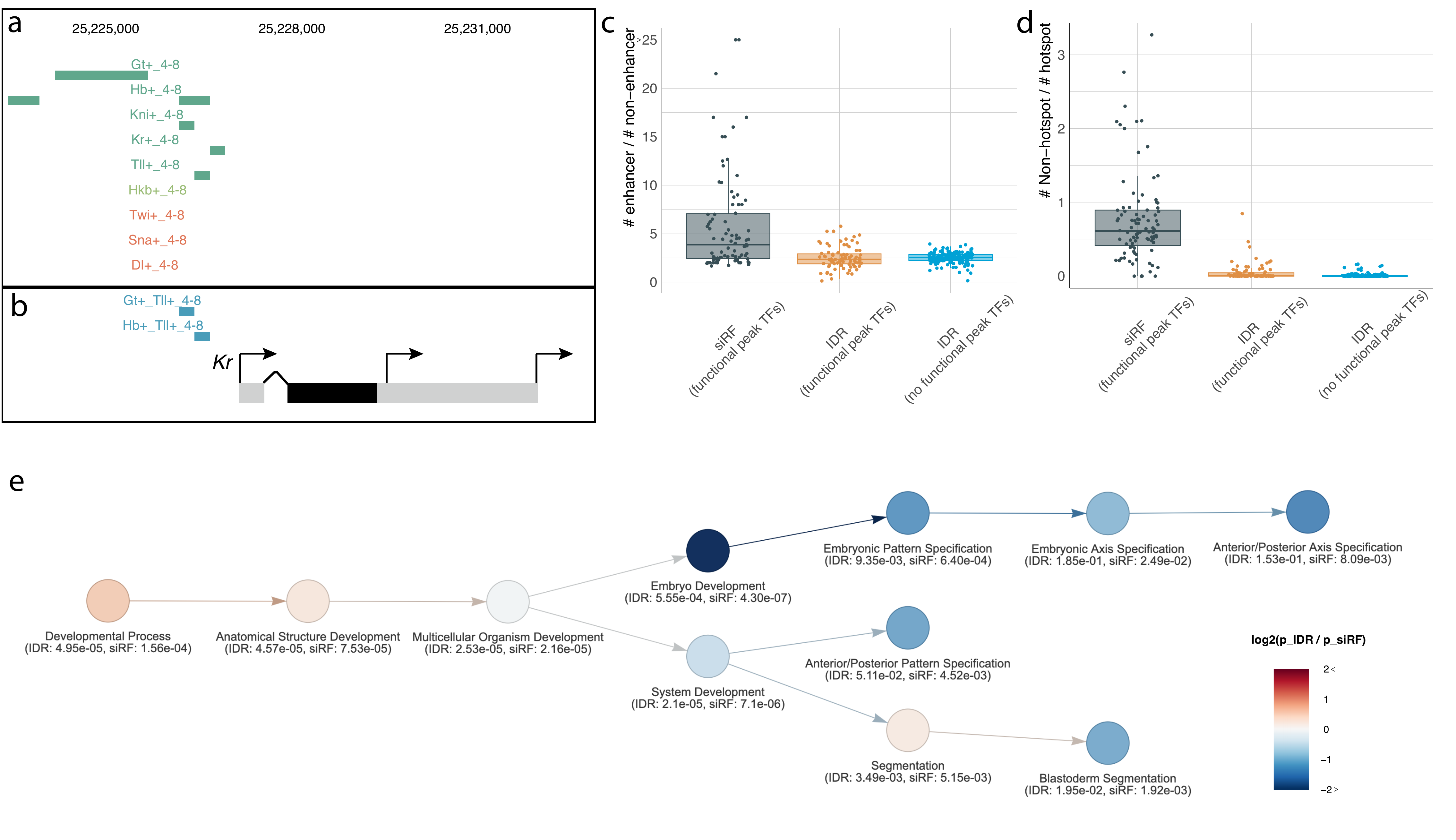}
  \end{center}
  \caption[Functional TF peaks estimated by siRF]{ Functional peaks estimated by siRF. \textbf{a-b.} Genome browser snapshot of siRF functional peaks surrounding Kr for gap-related and DV TFs (\textbf{a}) and signed interactions discovered by siRF (\textbf{b}).  color: dark green = gap TFs reported to regulate \textit{Kr}, light green = gap TF with no reported regulation of \textit{Kr}, red = DV TFs, blue = interaction peaks. \textbf{c} Ratio in the number of enhancer peaks / non-enhancer peaks. \textbf{d} Ratio in the number of non-hotspot peaks / hotspot peaks. points: TFs, color: grey = siRF-based functional peaks, yellow = IDR peaks for TFs with functional peaks called, blue = IDR peaks with no functional peaks called. \textbf{e}. GO enrichment analysis of genes based on IDR peaks or functional peaks for gap-related TFs (KR, KNI, TLL, HKB, HB, GT, CAD, BCD; gene defined as bound if peak within +/- 1kb from the genomic region). P-value derived from Fisher's exact test averaged across gap-related TFs. Tree hierarchy shows GO terms related to AP patterning and segmentation.
  } \label{fig:fig3}
\end{figure}

We used the signed interactions discovered by siRF to generate \textit{functional peak scores} at 550,274 genomic segments, covering the entire \textit{Drosophila} genome (Methods). These scores represent the predicted probability of enhancer activity at a genomic region as a function of interacting TFs. Functional peaks derived from these scores (Methods), each associated with a signed interaction, are thus predicted to lie within an enhancer element putatively regulated by constitutive TFs. To the best of our knowledge, this peak set represents the first genome-wide collection of putative TF binding events associated with enhancer activity. For example, functional peaks surrounding the widely-studied gap gene \textit{Kr} (Fig. \ref{fig:fig3}A,B), a gap gene that plays an important role in anterior posterior (AP) patterning, include known \textit{Kr} regulators Hunchback (HB)\cite{hoch1991gene}, Giant (GT)\cite{kraut1991mutually}, Tailless (TLL)\cite{steingrimsson1991dual}, Knirps (KNI)\cite{gaul1987pole} and KR itself\cite{warrior1990dose}.  Negative control TFs Huckebein (HKB), a gap gene that has no reported interaction with \textit{Kr} (for a review see \cite{jaeger2011gap}), and dorsal ventral (DV) patterning factors Twist (TWI), Snail (SNA), and Dorsal (DL) have no peaks surrounding \textit{Kr}. The full catalog of predicted enhancer elements, along with the individual TFs and TF-TF interactions our models predict drive their functions can be downloaded and explored as a \href{http://genome.ucsc.edu/cgi-bin/hgTracks?db=dm6&hubUrl=https://sina.lbl.gov/analysis/sirf_interactions/hub.txt}{UCSC genome browser track}. 

To systematically assess the quality of siRF-based functional peaks, we compared them with the modENCODE/modERN optimal IDR thresholded peak set\cite{celniker2009unlocking,kudron2018modern} (IDR peaks). Since siRF targets ChIP signal associated with enhancer activity,  we expected both fewer functional peaks than IDR peaks and enrichment of functional  peaks in enhancer regions. Consistent with our expectation, 169 / 262 TFs did not appear in any siRF interactions (i.e., the binding patterns of these eliminated TFs were not associated with enhancer activity). No functional peaks were called for these TFs (Supplementary Fig. S6a). Raw signal from these TFs had little predictive power for enhancer activity at any developmental stage or activity threshold (Supplementary Fig. S6b; Supplementary Table 4), and peaks were nearly all localized to regions previously annotated as hotspots\cite{kudron2018modern} (Supplementary Fig. S6c). For the remaining TFs, functional peaks showed an increase in the ratio of enhancer peaks to non-enhancer peaks compared with IDR peaks (mean ratio functional peaks = 6.62, mean ratio IDR peaks = 2.52, 2.6-fold increase; Fig. \ref{fig:fig3}c). Interestingly, by targeting enhancer-related peaks, siRF substantially reduced the number of peaks localized to hotspot regions; functional peaks showed an increase in the ratio of non-hotspot peaks to hotspot peaks (mean ratio functional peaks = 0.77, mean ratio IDR peaks = 0.052, 14.78-fold increase; Fig. \ref{fig:fig3}d). The biological relevance of these genomic hotspots is actively debated\cite{macarthur2009developmental,fisher2012dna,kasinathan2014high,siersbaek2014molecular,moorman2006hotspots}. Currently, studies of TF binding approaches remove these regions using ad hoc rules\cite{kudron2018modern}. Our results suggest that siRF can help highlight hotspot regions of potential biological interest and provide a systematic approach to removing noise.

We next investigated the ability of functional peaks to refine the signal of TF binding around a biological process of interest. Here we used functional peaks associated with gap-gene related TFs, whose role in regulating anterior/posterior (A/P) patterning and segmentation in the early embryo has been well established (for a review see \cite{jaeger2011gap}). GO terms related to A/P patterning and segmentation were more strongly enriched among genes neighboring functional peaks (peaks +/-1kb from the genomic region) compared with IDR peaks (Fig. \ref{fig:fig3}e-f). Indeed, GO terms most specific to A/P patterning consistently showed stronger enrichment in functional peaks compared with IDR peaks (Fig \ref{fig:fig3}e). These results suggest that among gap-related TFs, functional peaks show greater specificity for known biological functions.

Beyond peak calling, interactions discovered by siRF suggest interesting biological hypotheses. Here, we focused on \textit{caudal (cad)}, one of the earliest genes expressed in \textit{Drosophila} embryos\cite{moreno1999caudal}. Translation of the maternally deposited ubiquitous \textit{cad} transcripts is repressed at the anterior end by Bicoid (BCD) thus establishing a posterior gradient of CAD protein that is essential for development of the most posterior parts of the fly embryos. In total, siRF discovered 21 signed interactions surrounding CAD, all from stage 4-8 enhancer models (Supplementary Table 3), recapitulating and identifying new interactions that are expected to play an important role in early embryonic development. 

\begin{figure}
  \begin{center}
    \includegraphics[width=0.8\textwidth]{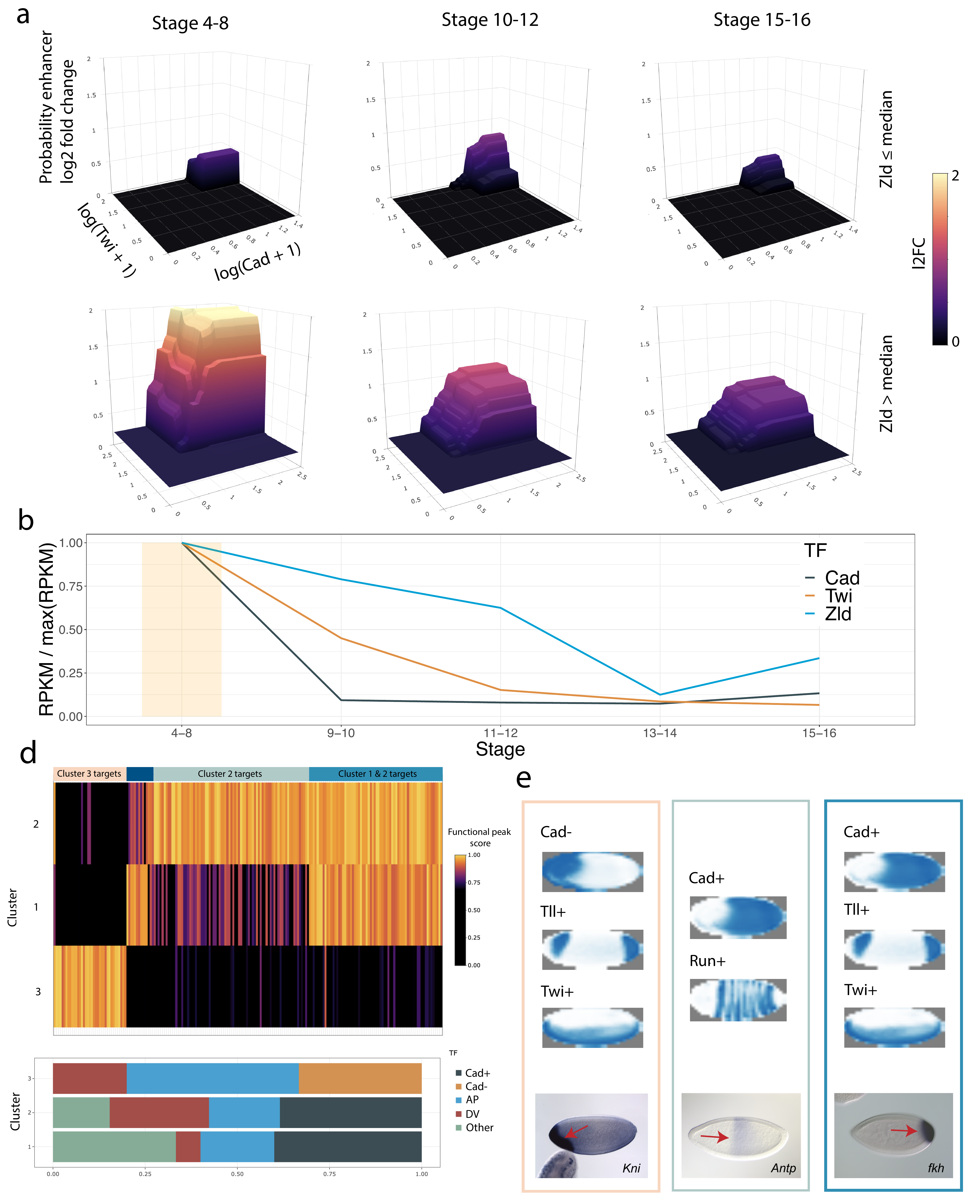}
  \end{center}
  \caption[Spatiotemporal dynamics of putative interactions]{Spatiotemporal dynamics of putative interactions. \textbf{a.} Response surfaces reporting the predicted probability log2 fold change (FC) of enhancer activity by stage (columns) as a function of Cad, Twi, and Zld binding levels. FC evaluated relative to average probability of enhancer activity for each stage. Top row = genomic segments where Zld is bound below median observed binding level, bottom row = genomic segments where Zld is bound above median observed binding level. Color = log2 FC. Response surfaces are thresholded below at 0 for visualization. \textbf{b.} RNA-seq time course data for Cad, Twi, and Zld expression, showing concordance with interaction activity as reported in response surfaces. \textbf{c.} Functional peak scores for interactions involving Cad, clustered into 3 groups. Color =  maximum functional peak score (across interactions in a cluster) at a genomic segment (columns), normalized to 0-1 scale. Columns are hierarchically clustered into groups representing putative targets of pink = cluster 3 interactions, dark blue = cluster 1 interactions, light blue cluster 2 interactions, blue = cluster 1 and 2 interactions. \textbf{d.} Proportion of TFs in cluster interactions that belong to different patterning groups AP = (Gt, Kr, Tll, Hkb, Hb, Kni, Run, Prd), DV = (Twi, Dl, Zen), Other = other TFs. \textbf{e.} Representative images of CRMs targeted by interactions in each cluster (bottom) and quantified spatial expression profiles of putatively interacting TFs taken from \cite{wu2016stability}. Color borders correspond to column clusters in \textbf{d}. Web apps for visualizing response surfaces and localized interaction prediction clusters are available through the \href{http://monster.lbl.gov:3838/sample-apps/}{BDGP}.} \label{fig:fig2}
\end{figure}

We assessed the activity of putative CAD interactions across embryonic development using stage-specific response surfaces (Methods), which report enrichment in enhancer activity as a function of TF binding (Fig. \ref{fig:fig2}). For example, regions bound by high levels of CAD, Twist (TWI), and Zelda (ZLD) show a 4-fold increase in the probability of stage 4-8 enhancer activity relative to regions where any one of those TFs is bound at low levels (Fig. \ref{fig:fig2}a). The effect size drops over the course of embryonic development in concordance with RNA-seq time course measurements of these three TFs (Fig. \ref{fig:fig2}b). Our results are consistent with the known roles of these factors in early embryonic development. CAD is highly expressed in ovaries, maternally deposited, and zygotically expressed at much lower levels—2-4 hrs (half the maternal level) and the rest of embryogenesis ($\sim1/10^{th}$ the maternal level). ZLD is also highly expressed in ovaries and maternally deposited. Its highest expression is at 2-4 hrs in early blastoderm as it plays an important role as a zygotic genome activator \cite{liang2008zinc}. TWI is not maternally deposited, though its highest level of expression is also 2-4 hrs, consistent with its role in gastrulation \cite{simpson1983maternal}. 

To compare the activity of CAD interactions at different genomic locations, we clustered the 21 CAD interactions based on their functional peak scores at enhancers. Interactions clustered into three groups broadly defined by (i) low levels of CAD binding (ii) high levels of CAD binding with other anterior-posterior (AP) TFs and (iii) high levels of CAD binding with dorsal-ventral (DV) TFs (Fig. \ref{fig:fig2}c-d; Methods). Spatial expression patterns associated with enhancers in each cluster—evaluated in whole mount embryonic reporter assays\cite{hammonds2013spatial}—were concordant with patterns implied by signed interactions (Fig. \ref{fig:fig2}e), highlighting the power of signed interactions to recapitulate the combinatorial logic that underpins organ differentiation and cell fate specification. An interactive web application for visualizing response surfaces and clustering interactions is available through the \href{http://monster.lbl.gov:3838/sample-apps/}{Berkeley \textit{Drosophila} Genome Project (BDGP)}. 

This study introduces siRF, a general-purpose, interpretable machine learning method that we used to annotate TF binding sites using ChIP data and a library of validated enhancer elements. Our strategy reframes the traditional task of peak calling—a form of unsupervised learning—to a supervised learning task. Indeed, each peak annotated using our approach is predicted to lie in an active enhancer element. On held-out test data, our method validates with remarkable accuracy, consistent with our prior results applying RF-based methods to enhancer prediction tasks\cite{arbel2019exploiting, basu2018iterative}. Using an extensive ChIP-seq/ChIP-ChIP dataset generated by the modENCODE, MODERN and BDTNP Consortia, and an RNA-seq dataset generated by the modENCODE Consortium, we showed that our predicted interactions conform well with prior knowledge at well-studied enhancer elements. Further, we found that predicted interactions recapitulate expression dynamics measured by RNA-seq data in time course—a particularly intriguing observation, since RNA-seq time-course information in no way figured into the training of the model. Unsurprisingly, the resulting library of functional peaks was heavily enriched for known enhancer elements compared to traditional IDR peaks, and was depleted, for most TFs, for localization to ``hotspots'', which are widely regarded as artifacts—though there is some debate about this\cite{macarthur2009developmental,fisher2012dna,kasinathan2014high,siersbaek2014molecular,moorman2006hotspots}. Integrative analysis of enhancer and ChIP-seq data provides a unique opportunity to learn both functional binding events and interactions between TFs that give rise to patterned gene expression. siRF reveals strong statistical associations between groups of TFs and enhancer activation that drives the spatiotemporal landscape of gene expression during development. Further work will be needed to determine if the correlations detected by siRF are substantiated by direct physical interactions between TFs and the genes they control.

\section*{Methods}
\section{Signed iterative random forests}
We developed a two-phase approach to detect enhancer-related, functional TF peaks. First, we use a training dataset of annotated enhancers to search for statistical interactions among TFs that are associated with enhancer activity. We refer to this first phase as \textit{interaction discovery}. The interaction discovery phase substantially reduces the set of $O(p^s)$ possible order-$s$ interactions among $p$ TFs. This makes it computationally tractable to map putative interactions across the genome. In the second phase, we compute genome-wide, functional peak scores for interactions identified in the interaction discovery phase. These scores measure the similarity between TF binding patterns at a specific genomic location and the binding patterns associated with a given interaction. We refer to this second phase as \textit{interaction localization}.

The signed iterative random forest algorithm (siRF) unifies interaction discovery and localization. Although we developed siRF for the purposes of mapping TF interactions, it is a general purpose machine learning algorithm to (i) detect high-order, rule-based feature interactions and (ii) generate predictive rules associated with these interactions. In the following sections, we motivate siRF in the context of the TF/enhancer problem considered in this paper and present the details of the algorithm in its general form.
\subsection{Interaction discovery through signed iterative random forests (siRF)}\label{sec:sirf-method} 
Data driven discovery of TF interactions at enhancer elements amounts to learning relationships of the form ``TF1, TF3 are unusually enriched and TF2, TF4 are unusually depleted in many known enhancer locations.''  We introduce \textit{signed interactions}, extracted from decision tree ensembles, to quantitatively encode and evaluate such binding patterns. Signed interactions represent the previous statement as the set $\{+1, -2, +3, -4\}$ (where $+/-$ indicate enrichment/depletion and values TF indices), allowing us to address two important aspects of the statement above: (i) what constitutes unusual enrichment or depletion of a TF? (ii) which signed interactions are active at genomic regions of interest (e.g. many known enhancer locations)?

We seek to discover signed interactions through a supervised learning framework. Consider data $\{ (y_i, \mathbf{x}_i)\}_{i=1}^n$, where $y_i \in \{0, 1\}$ denotes a binary label and $\mathbf{x}_i = (x_{i1}, \ldots, x_{ip}) \in \mathbb{R}^p$ a vector of $p$ continuous features. In the case of enhancer prediction, samples $i$ represent genomic segments, labels $y_i$ whether the $i^{th}$ segment is an active enhancer or not, and feature vectors $\mathbf{x}_i$ continuous measurements (ChIP-chip or ChIP-seq) on $p$ different TFs. An ensemble of decision trees (e.g. an RF) can be viewed as a collection of decision rules, each corresponding to a single leaf node in a single tree, of the form
$$
\mbox{if $x_1 > t_1, \, x_2 < t_2, \ldots, x_k > t_k$, then $y = 1$}.
$$
The iRF algorithm \cite{basu2018iterative} encodes such a rule in the form of an \textit{unsigned interaction} $\{1, 2, \ldots, k \}$, representing the combination of TFs that appear together in a rule. To encourage more stable use of unsigned interactions on RF decision paths, iRF fits iteratively re-weighted RFs—weighting feature selection by mean decrease in impurity (MDI) feature importance from the previous iteration.  It then uses random intersection trees (RIT) \cite{shah2014random} to search for frequently occurring unsigned interactions associated with enhancers (i.e. $y = 1$). 

In siRF, we refine this search to distinguish between feature enrichment and depletion. Re-writing the unsigned interaction above as the signed interaction $\{+1, -2, \ldots, +k \}$ captures both selected features and the direction of their associated inequalities (\sref{sec:signed}). For every observation $i$, we generate $T$ (number of trees in the forest) encodings of the form
$$
(S_{i_t}, Z_{i_t}),\,  t = 1, \ldots, T,
$$
where $S_{i_t} \subseteq \{ \pm 1, \pm 2, \ldots, \pm p \}$ represents the signed interaction associated with the leaf node in tree $t$ that contains observation $i$, and $Z_{i_t} \in \{0, 1\}$ the prediction of that leaf node. This can be interpreted as a list of putative interactions, offered by $T$ different ``experts'' (trees in a forest). In the context of enhancer prediction, this list represents the patterns of TF enrichment/depletion at genome location $i$ that lead to predicted enhancer status $Z_{i_t}$. As in iRF, we use RIT to search for frequently occurring signed interactions from the space of all $nT$ encodings (\sref{sec:rit}).

\subsubsection{Signed interactions}\label{sec:signed} 
We define feature enrichment and depletion through a mapping $\mathbf{x} \mapsto S \subset \{\pm 1, \pm 2, ..., \pm p\}$ based on RFs. A feature is deemed enriched (resp. depleted) in decision rules that rely on high (resp. low) levels of the feature for prediction. Our encoding is similar to \cite{dinalankara2018digitizing}, who binarize omics profiles based on divergence from a reference distribution. However, rather than defining activity relative to a reference distribution, we define enriched/depleted features through decision paths of a RF (or variant thereof). By building on Boolean rules that model thresholded feature levels, our encoding incorporates both combinatorial and thresholding phenomena known to underlie important biomolecular interactions \cite{wolpert1969positional}.

Consider a decision tree node that splits on feature $j\in\{1,\dots,p\}$ with threshold
$t_j\in\mathbb{R}$. Observations $\mathbf{x}$ that arrive at this node will be
sent to the left child when $x_j < t_j$ and be sent to the right child
otherwise. We use the signed feature-index $\gamma \in \{ \pm 1, \pm 2, ..., \pm p \}$ to
describe the selected splitting feature and inequality direction associated with
a child node. The left child is represented by the signed feature index
$-j$ and the right child by $j$
(Fig. S1). Thus, signed feature indices describe whether a
node is associated with relatively high or low levels of the selected feature. We 
define the signed interaction associated with leaf node $l=1,\dots,L$
as 
\begin{equation}
  S_l = \{s_1, \dots, s_k\}\subseteq\{ \pm 1, \pm 2, ..., \pm p \},
\label{eq:signed}
\end{equation}
where $s_1, \dots, s_k$ are the signed feature indices corresponding
to all non-root nodes along the decision path. For
simplicity, we assume that each feature is selected at most once along a
decision path. We discuss the case where a feature is selected multiple times on a decision path in \sref{sec:redundant}.

Signed interactions define decision paths in an RF up to threshold values $t_j$. We say that a leaf node $l$ contains a signed interaction $S$ if $S\subseteq S_l$.  This allows us to relate paths that use the same features in similar ways, but whose thresholds vary as a result of the randomness inherent to RF. Individual decision paths that contain the same signed interaction $S$ can be grouped into a collection of ``similar'' rules that capture the same pattern of enrichment/depletion among select features but use slightly different thresholds.

\subsubsection{Searching for signed interactions through iRF} \label{sec:rit}
We search for signed interactions that frequently appear on RF decision paths using generalized random intersection trees (gRIT) \cite{basu2018iterative}. For every leaf node in an iteratively re-weighted RF, we generate signed interactions and encode each observation/tree pair as

\begin{equation*}
(S_{i_t}, Z_{i_t}) \hspace{1em} i=1,\dots,n, \hspace{1em} t=1,\dots,T,
\end{equation*}
where $S_{i_t}$ represents the signed interaction associated with the leaf node
containing observation $i=1,\dots,n$ in tree $t=1,\dots,T$ and $Z_{i_t}$ the corresponding
prediction. These pairs are used as inputs to RIT, a randomized search algorithm that detects frequently co-occuring feature combinations—in our case signed interactions—from the space of all $nT$ encodings \cite{shah2014random}.   

\subsubsection{Evaluating signed interactions via stable and predictive importance metrics}\label{sec:imp}
We view signed interactions as candidate regulatory interactions. The standard RF importance metrics (e.g. mean decrease in Gini Impurity, mean decrease in accuracy) have been widely used to evaluate the contribution of individual features to model predictions. However, these metrics do not evaluate the contribution of interactions to model predictions. While permutation-based approaches have been proposed to evaluate pairwise interactions in RFs \cite{li2016detecting}, they have not been generalized to higher-order interactions. Moreover, permutation strategies can be misleading as they often depend on model estimates in regions of the feature distribution far from observed training data \cite{hooker2021unrestricted}. 

In this section, we describe two importance metrics: precision and prevalence for evaluating/ranking signed interactions directly from an RF (Fig. S2). For each metric, we define null models that serve as controls motivated by properties relevant to enhancer activity. At a high level the null metrics are inspired by \cite{elsayed2017structure}, who consider the problem of generating controls for single neuron data.

\textbf{Prevalence describes the stability of a signed interaction along decision paths.} We define the \textit{prevalence} of a signed interaction $S$ for a class $C\in\{0,1\}$ relative to a fitted RF as
\begin{equation}
  P(S|C) := \frac{1}{T}\cdot
  \sum_{t=1}^{T}
  \frac{\sum_{i=1}^{n} 
      \mathds{1}(S\subseteq S_{i_t}) \cdot 
      \mathds{1}(y_{i} = C)}
    {\sum_{i=1}^{n} \mathds{1}(y_{i} = C)}.
\label{eq:prevalence}
\end{equation}
Prevalence indicates the proportion of class-$C$ observations for which an interaction $S$ is used in predictions from the fitted RF. It represents a notion of interaction stability—specifically stability relative to the data and model perturbations used to construct each tree. Here, stability is distinct from the stability scores proposed by \cite{basu2018iterative}, who evaluate stability relative to ``outer layer'' bootstrap samples. Prevalence, on the other hand, evaluates stability relative to a single run of RF (or variant thereof) and can be paired with outer layer bootstrap sampling to assess stability over multiple runs. 

\textbf{Feature selection dependence ($FSD$) evaluates interaction effects relative to prevalence}: Cooperative binding among TFs leads to nonlinear relationships between TF concentration and binding affinity. Such binding activity plays an important role in producing ``switch-like'' effects that typify enhancers \cite{spitz2012transcription}. From a biological perspective, it is therefore valuable to identify TFs whose binding patterns depend on the presence/absence of other TFs. We evaluate such dependencies through feature (i.e. TF) co-occurance rates along RF decision paths relative to the null hypothesis
\begin{equation}
  H_0^{(FSD)} : P(S|C) = \prod_{s\in S} P(s|C).
  \label{eq:h0fsd}
\end{equation}
Intuitively, equation \eqref{eq:h0fsd} states that the individual features in a signed interaction appear independently of one another along decision paths of a fitted RF (Fig. S3). In the context of enhancer prediction, $H_0^{(FSD)}$ states that the presence of a given TF in $S$ does not depend on the presence/absence of other TFs in $S$. To quantify deviation from this null hypothesis, we define the \textit{feature selection dependence} ($FSD$):
\begin{equation}
  FSD(S;C) := \log{\left(\frac{P(S|C)}{\prod_{s\in S} P(s|C)}\right)}
  \label{eq:intstr}.
\end{equation}
Equation \eqref{eq:intstr} compares the prevalence of a signed interaction $S$ with its expected prevalence if features were selected independently of one another. In a broad range of simulations (3 Boolean models defined over both simulated and real data), we find that filtering interactions based on $FSD$ helps differentiate between additive and non-additive components of a generative model (Fig. S7).

\textbf{Precision describes how accurately a signed interaction predicts enhancer activity.} We define the \textit{precision} of a signed interaction $S$ for a class $C$ relative to a fitted RF as
\begin{equation}
  P(C|S) := \frac{1}{T}\cdot
  \sum_{t=1}^{T}
  \frac{\sum_{i=1}^{n} \mathds{1}(S\subseteq S_{i_t}) \cdot \mathds{1}(y_i = C)} 
    {\sum_{i=1}^{n} \mathds{1}(S\subseteq S_{i_t})},
\label{eq:accuracy}
\end{equation}
where $S_{i_t}$ is the signed interaction associated with the leaf node in tree $t$ containing observation $i$. Precision measures the average proportion of observations that belong to a given class among leaf nodes in a RF (i.e. decision paths) that contain a signed interaction $S$. We evaluate precision on out of bag (OOB) training samples rather than held-out testing data. This allows us to evaluate precision directly from a fitted RF and use testing data for down-stream analysis of selected interactions. When data are heterogeneous, precision can highlight signed interactions that achieve accurate prediction for select subpopulations, even if the full RF is a weak predictor across the full dataset (Fig. S2).

\textbf{Minimum increase in precision ($MIP$) measures the contribution of additional features to interaction precision}: 
TFs bind many inactive regions in the genome \cite{li2008transcription}, suggesting that binding alone is not indicative of functional regulation. To distinguish putative interactions where all TFs play a functional role from those where some subset of TFs are binding in a non-functional manner, we define a hypothesis testing framework that quantifies functional/non-functional binding relative to changes in prediction accuracy. Specifically, we define the null hypothesis

\begin{equation}
  H^{(MIP)}_0 : P(C|S) = \max_{S'\subset S} P(C|S'),
  \label{eq:accdiff}
\end{equation}
which states that for a putative interaction $S$, there is some $S' \subset S$ (i.e. $S' \neq S$) that achieves comparable precision. In the case of enhancer prediction, this hypothesis states that only a subset of the TF binding events implied by $S$ are required to predict enhancer activity at the observed level of precision. To quantify deviation from this null hypothesis, we define the \textit{minimum increase in precision} ($MIP$) as

\begin{equation}
  MIP(S;C) := \min_{S'\subset S: |S'| = |S| - 1}  \log{\left(\frac{P(C|S)}{P(C|S')}\right)},
  \label{eq:accdiff}
\end{equation}
Equation \eqref{eq:accdiff} compares the precision of a signed interaction to
that of its most precise subset (Fig. S3a). Note, in equation \eqref{eq:accdiff} we restrict ourselves to subsets $S'$ with one fewer element than $S$ to avoid an exponential increase in the number of terms that must be evaluated. Smaller subsets generally achieve comparable or worse precision than those considered in \eqref{eq:accdiff} since predictions leverage information from fewer features. In simulation studies, we find that filtering based on MIP removes interactions that include both ``active'' features (i.e. those used to generate a response) and ``inactive'' noise features (Fig. S7).

\subsubsection{PCS filtering to identify candidate TF interactions}\label{sec:pcs}
We filter interactions discovered by siRF using the recently proposed PCS framework \cite{yu2019three}. The PCS framework builds on three fundamental principles of data science: predictability, computability, and stability. It unifies and expands on ideas from statistics and machine learning, using predictability to evaluate the empirical evidence for a model, computational strategies to simulate realistic reference/null distributions, and stability to assess the reproducibility of findings. In the setting of TF interaction discovery, precision (relative to enhancer prediction) measures of predictive accuracy associated with each signed interaction. FSD and MIP provide computationally-derived models to evaluate signed interactions against biologically motivated null hypotheses. Finally, we evaluate the stability of signed interactions relative to an ``outer layer'' bootstrap sample of the training set.

First, we generate a set of candidate signed interactions from iteratively re-weighted RFs trained on the full set of training samples for $K=5$ iterations. We then train a collection of $B=50$ feature-weighted RFs on bootstrap samples from the training set, setting feature sampling weights equal to those used in the final iteration. This results in a set of $50$ distinct iteratively re-weighted RFs (i.e. derived from 50 bootstrap samples of the training data) each trained using the same feature sampling probabilities. For all signed interactions recovered in the first step, we evaluate the average: precision, prevalence, FSD, and MIP across each of the 50 bootstrap RFs. In addition, we compute stability scores for (i) the proportion of times (over bootstrap RFs) an interaction is recovered by RIT (as in \cite{basu2018iterative}), denoted as $sta(S)$ and (ii) the proportion of times (across bootstrap RFs) $FSD(S;C) > 0$ and $MIP(S;C) > 0$, denoted as $sta(S; MIP)$ and $sta(S; FSD)$ respectively. Given these stability scores, we filter interactions as follows, using $0.5$ as a stability threshold as in \cite{basu2018iterative}:

\begin{enumerate}
  \item \textbf{Prediction screening:} We remove any interaction for which $sta(S;MIP) < 0.5$. 
  \item \textbf{Stability screening:} We remove any interaction with $sta(S)< 0.5$.
  \item \textbf{Null model screening:} We remove any interaction for which $sta(S;FSD) < 0.5$. 
  \item \textbf{Concordance screening:} We remove all interactions where TF expression, as measured from an independent dataset of RNA-seq time course expression (see \sref{sec:concordance}), was not concordant with the enhancer label embryonic developmental stage where the interaction was recovered.
\end{enumerate}
The combination of filtering in steps 1-3 consistently recovered true data generating interactions across simulation studies representing a range of biologically-inspired, Boolean-like generating models (Fig. S7). Step 4 adds an additional biological constraint, ensuring that recovered interactions correspond to combinations of TFs are co-expressed in overlapping time windows.

\subsection{Interaction localization to compute functional peak scores}
\label{sec:prediction}
The interaction discovery process described above generates a set of candidate interactions. Here, we seek to identify genomic regions where TF binding patterns resemble putative interactions. Toward this end, we decompose RF decision rules into modules associated with discovered signed interactions. Each module is a predictive model that we use to score the degree to which TF binding patterns at unobserved genomic segments resemble signed interactions discovered from the training data. We refer to these scores as \textit{functional peak scores}.

Consider an RF, or variant thereof such as iRF, with $T$ decision trees and leaf nodes $l=1,\dots,L$ across all trees in the RF. Each leaf node $l$ is associated with (i) a class label $Z_l\in \{0,1\}$, (ii) a region of the feature space $\mathcal{R}_l=(R_1, \dots, R_p)\subseteq\mathbb{R} ^ p$, and (iii) a signed interaction $S_l$. We note that leaf node predictions $Z_l$ depend on the distribution of training sample responses falling in that node, though we omit this dependency for simplicity. The RF prediction $\hat{y}(\mathbf{x})$ for an observation $\mathbf{x}\in\mathbb{R} ^p$ can be written as

\begin{equation}
\hat{y}(\mathbf{x}) = \frac{1}{T} \sum_{l = 1} ^ L Z_l \cdot r(\mathbf{x}, \mathcal{R}_l, S_l), \;\;\mbox{where} \hspace{1em} r(\mathbf{x};\mathcal{R}, S) := \prod_{j \in \{|k|:k\in S\}} \mathds{1}(x_j \in R_j), 
\label{eq:rfruleprediction}
\end{equation}
and $|\cdot|$ denotes absolute value. Here, the denominator $T$ comes from the fact that each observation $\mathbf{x}$ is active in exactly one rule (i.e. falls in exactly one leaf node) per tree while $|\cdot|$ is used to reference features in a signed interaction without the sign component.

Functional peak scores modify \eqref{eq:rfruleprediction} to define the RF prediction associated with a signed interaction $S$ as

\begin{equation}
\hat{y}(\mathbf{x};S) = \frac{1}{|\{l:S\subseteq S_l\}|} \sum_{l:S\subseteq S_l} Z_l \cdot r(\mathbf{x}, \mathcal{R}_l, S).
\label{eq:sirfprediction}
\end{equation}
There are two key differences between \eqref{eq:rfruleprediction} and \eqref{eq:sirfprediction}. First, \eqref{eq:sirfprediction} takes an average over leaf nodes containing a signed interaction while \eqref{eq:rfruleprediction} averages over all leaf nodes in an RF. Second, \eqref{eq:sirfprediction} is restricted to the portion of the decision rule corresponding to $S$. As a result, \eqref{eq:sirfprediction} represents an average over rules $r$ with similar—i.e. using the same features and inequality directions—thresholding behavior. Predictions of this form can be viewed as a decision rule with smooth decision boundaries, defined over the features in $S$. Such predictions exhibit several desirable properties: 

\begin{enumerate}

\item Smooth decision boundaries make predictions less sensitive to small
  changes in $\mathbf{x}$ compared with a single decision rule defined over the
  same features. 
  
\item Predictions incorporate non-linear feature interactions in the form of rules as in equation \eqref{eq:rfruleprediction}.

\item Predictions depend solely on the set of interacting features $\{|j|:j\in S\}$, and are therefore sparser than the full decision tree ensemble. We note that sparsity has been advocated as an important consideration for interpretable machine learning \cite{murdoch2019interpretable}.

\item Predictions are built from a collection of rules that define similar—i.e. using the same features and inequality directions—functional relationship with responses, which can be summarized in interpreted through $S$.

\item Predictions are monotonic as a function of each active feature, and
  the sign of each $j\in S$ defines whether predictions are increasing or
  decreasing as a function of the corresponding feature.

\end{enumerate} 

\subsubsection{Generating genome-wide siRF functional ChIP peaks}
We generated functional peaks for each interaction and all individual TFs involved in discovered interactions. For interactions discovered in multiple developmental stages, we generated one set of peaks per stage in which the interaction was discovered. These peaks were generated as follows.
\begin{enumerate}
\item Functional peak scores \eqref{eq:sirfprediction} were evaluated at $n=550274$, 500bp segments spanning the \textit{Drosophila} genome (250 bp overlap). Rules in \eqref{eq:sirfprediction} were derived from the RF models trained for the stage and activity threshold in which the interaction was discovered.
\item genome-wide functional peak scores were smoothed using a uniform kernel with a 2000bp bandwidth.
\item Within each chromosome, smoothed functional peak score signatures were cut into contiguous blocks at local minima.
\item Each contiguous block was trimmed to peaks representing segments with the maximal functional peak score within that block.
\item Peaks were filtered based on their maximal functional peak scores. Thresholds for filtering were set to achieve a false positive rate of $0.001$, estimated from hold-out test samples of the annotated enhancer dataset. We note that the proportion of enhancers in this dataset is not representative of the entire \textit{Drosophila} genome, and thus $0.001$ does not represent our expected false positive rate for genome-wide peaks (Supplementary Table 5). 
\item For single TF peaks only, we removed peaks where an interaction peak (involving the TF of interest) was not present. I.e., all single TF peaks appear in regions with at least one interaction peak involving that TF. 
\end{enumerate}

\subsubsection{Response surfaces}
\label{sec:surface}
Functional peak scores can be visualized as \textit{response surfaces} that report predicted response values as a function of interacting features. Response surfaces provide a qualitative depiction of what a RF ``sees'' with respect to a given interaction. We generate response surfaces by evaluating functional peak scores across a uniformly spaced grid of points to map out predictions across the entire feature space. To increase computational efficiency, we exploit the fact that predictions from a single rule are constant within the region $\mathcal{R}$ associated with that rule. This allows us to evaluate a single point within each region rather than multiple points across the grid. Intuitively, our representation uses each region $\mathcal{R}_L:S\subseteq S_l$ as a data-adaptive binning to generate a histogram of response values. Averaging over multiple regions allows us to identify the smoothed boundaries learned by an RF. 

To assess variation in response surface activity across subpopulations, we replace $Z$ in \eqref{eq:rfruleprediction} with the subpopulation's average response in each region. Specifically, let $\mathcal{I}\subset \{1,\dots,n\}$ index a subpopulation of interest and $\mathcal{I}_l = \{i \in \mathcal{I} : \mathbf{x}_i \in \mathcal{R}_l\}$ the samples in $\mathcal{I}$ that fall in leaf node $l$. Response surfaces for a subpopulation are then generated from predictions
\begin{equation}
    \hat{Z}_l = \frac{1}{|\mathcal{I}_l|} \sum_{i\in \mathcal{I}_l} y_i.
\end{equation}
We use response surfaces targeted at subpopulations to (i) assess generalizability an interaction by reporting response surfaces generated on held-out test data and (ii) assess how interaction activity changes over time by reporting response surfaces generated on the subset of samples that are active at each developmental stage. 

\subsubsection{Activity profiles from functional peak scores}
Signed interactions can be related based on their functional peak scores evalauted over a set of genomic locations of interest. Intuitively, this asks whether signed interactions are active at similar regions of the genome. For a set of regions $i=1,\dots,R$ and signed interactions $S_1, \dots, S_K$, we generated functional peak score profiles $(\hat{y}(\mathbf{x}_1;S_k), \dots, \hat{y}(\mathbf{x}_R;S_k)) \in \mathbb{R} ^ R$, $k=1\dots, K$, measuring the activity of each signed interaction at each region. Clustering interactions based on these profiles allowed us to group signed interactions that exhibit similar behavior across the target regions of interest (e.g., \href{http://monster.lbl.gov:3838/sample-apps/}{Berkeley \textit{Drosophila} Genome Project (BDGP) interaction clustering}). 
\section{Redundant feature selection on decision paths}\label{sec:redundant}


In some cases, particularly for deep decision trees used by RF, a feature $j$
may be selected multiple times on the same decision path. In these situations
both signed feature indices $j$ and $-j$ could appear on the decision path.  The
recursive nature of decision tree partitions suggests a natural way to handle
this issue.  For any features that are selected multiple times on a decision
path, we use only the signed feature index associated with the first split,
since subsequent splits on the same feature are restricted by the initial split. 
This representation destroys the one-to-one correspondence between a signed
interaction and feature splits along a decision path. However, we find
empirically that our simplified representation performs comparably to the full
decision rule in terms of predictive accuracy and allows for a high-level,
interpretable grouping of decision rules.

\section{Data processing and modeling}\label{sec:enhancer_sirf}
\textbf{Enhancer labels:} Enhancer labels (i.e. responses) were taken
from the \href{http://enhancers.starklab.org/search}{Stark Lab database of fly
enhancers} \cite{kvon2014genome}. This database contains labels of enhancer
activity for 7705 sequences, representing 13.5\% of the non-coding,
non-repetitive \textit{Drosophila} genome. Labels were generated using a
high-throughput, stable-integration, transgenic assay. Briefly, a genomic
sequence (100-3000nt) was placed into a reporter construct and integrated into a
targeted site in the genome. The transgenic fly line was then amplified, embryos
were collected, fixed, and stained using immunohistochemistry to detect the
reporter \cite{tautz1989non, weiszmann2009determination}. The stained fly line
was then imaged and (i) manually annotated using a controlled vocabulary (CV)
\cite{tomancak2002systematic, tomancak2007global} to determine where and when
expression is driven and (ii) scored on a scale from from 1 = very weak to 5 = very 
strong based on the intensity of the corresponding expression pattern. 

We used these annotations to define a binary classification problem: predicting which 
sequences represented active enhancers different stages of embryonic development. To 
identify interactions with broadly defined enhancer activity, rather than activity associated 
with specific CV terms, we computed the maximum score across all CV terms for each 
sequence and stage window. To convert scores to binary responses, we considered 3 
different thresholds (2, 3, 4). Any sequence with score greater than or equal
to a given threshold at a specific stage window was considered an active enhancer at that 
threshold/stage. Any sequence with score = 0 over all stage windows was considered 
an inactive enhancer. Following the original paper \cite{kvon2014genome}, we did not consider 
a score of 1 (very weak enhancers) as active enhancers in our analysis. In 
addition, we did not consider scores of 5 (very strong enhancer) in our analysis due to a 
limited number of sequences annotated at this level. Based on these definitions, the number of 
observations used in our models varied (depending on stage/strength ranking threshold)  from 4398 
- 6862.

\textbf{DNA binding:} ChIP-seq experiments available through the \href{https://www.encodeproject.org/}{modENCODE/modERN} consortia and \href{http://bdtnp.lbl.gov/Fly-Net/}{Berkeley \textit{Drosophila} Transcription Network Project (BDTNP)} provide genome-wide, quantitative maps of DNA binding for a substantial portion of TFs in the \textit{Dosophila} embryo \cite{kudron2018modern,macarthur2009developmental,li2014establishment}. In total, we considered 316 experiments measuring DNA binding activity of 262 unique TFs. Segments used in our analysis ranged from $\sim$100-3500bp, with the majority ($\sim$80\%) ranging from 
2000-2500bp. To quantify binding levels for a segment/TF pair, we computed the maximum normalized fold enrichment of the given TF over the given segment. We note that this approach uses the raw ChIP signal rather than signal processed through peak calling methods such as irreproducible discovery rate (IDR). While our approach is likely to generate noisy scores for each segment/TF pair, it also maintains signal that can be lost through preprocessing. We used the raw measurements as this allowed siRF to determine which signal is important for a given prediction problem.  Moreover, we found that models trained on IDR peaks had generally poor predictive performance. We note that most TFs were assayed over a single time window, selected based on RNA-seq measurements. However, a small handful of TFs were measured at multiple time points during embryonic development. We treat TF replicates measured at different time points as distinct features, allowing siRF to select which replicate best predicts enhancer activity at a given developmental stage. 

\textbf{Training/test split:} Studies of the 3D architecture of the genome are
beginning to shed light on spatial dependencies that play an important role in
gene regulation. Ideally, we would like our training/testing split to respect
these dependencies. However, current knowledge of the 3D structure of the genome
does not suggest a well-justified approach for sample splitting. Since we were
interested in identifying candidate interactions across the entire genome, we
randomly split sequences into equally sized training and test sets. All reported
measures of prediction accuracy and response surfaces were evaluated on held-out
test data. Training/test splits are available in supplementary data files for each 
model.

\textbf{siRF parameters:} To predict enhancer
activity, we ran iRF for the default $5$ iterations with $50$ bootstrap samples and $1000$ trees for each iteration. 
We set RIT parameters to $n_{tree} = 5000$, $n_{child} = 2$ and $d = 2$, resulting in less stringent interaction 
filtering (i.e. returns more interactions) than the default values. This allowed us to 
filter interactions based on criteria relevant to enhancer prediction rather than relying 
on RIT for interaction filtering.

\textbf{Feature subsampling:} Using siRF to model TF binding introduces the potential for ``masking'' effects — i.e. when a particular feature is selected on a decision path, correlated features tend not to appear on the path. As a result, siRF can ``miss'' important TFs that are highly correlated with others. To mitigate this issue, we used a feature subsampling strategy. Specifically, we randomly sampled $50\%$ of TFs and ran siRF using only the selected subsample. We repeated this procedure 100 times and averaged importance metrics for each interaction across each replicate where it was recovered.

\subsection{Concordance filtering with temporal expression data}\label{sec:concordance}
The analyses described in this paper model enhancer activity across 5 time windows throughout \textit{Drosophila} embryonic development: stages 4-8 ($\sim$90-230 minutes post fertilization), 9-10 ($\sim$230-320 minutes post fertilization), 11-12 ($\sim$320-580 minutes post fertilization), 13-14 ($\sim$580-680 minutes post fertilization), 15-16 ($\sim$680-900 minutes post fertilization). At each time window, we extracted a collection of candidate TF interactions (signed interactions) and evaluated whether the TFs involved could plausibly interact using an independent RNA-seq time course dataset. For example, we filtered interactions identified in our stage 4-8 enhancer models to include only those among TFs expressed during stages 4-8, as determined by RNA-seq data \cite{graveley2011developmental}.

For a given interaction and stage, we computed the average reads per kilo base per million mapped reads (RPKM) across all putatively interacting TFs in the select stage. We compared this value to the average RPKM across the same set of TFs over the remaining stages. That is, we took the difference in average (over interacting TFs) RPKM between a selected stage and all other stages. To prevent a single TF within a candidate interaction from dominating this difference, we normalized RPKM for each TF to the 0-1 scale prior to averaging.

The observed difference above represents the extent to which gene expression for a collection of TFs is enriched in a select developmental stage relative to all other developmental stages. To evaluate the magnitude of this difference, we generated null distribution by permuting gene labels in the RNA-seq time course data. Specifically, each gene in the RNA-seq time course data is associated with a profile measuring expression across multiple time windows. We permuted the gene labels associated with these profiles 100 times, preserving the temporal associations within a profile but destroying the association between particular genes and the stages it is expressed. This null distribution corresponds to the null hypothesis that expression for a set of interacting TFs shows no association with developmental stage. We removed interactions for which the observed difference was smaller than the null difference more than 5\% of the time.

\section{Simulations} \label{sec:sims}
To evaluate the signed interactions recovered by siRF, we developed a suite of
simulation experiments based on Boolean-type rules intended to reflect
stereospecific biological interactions. Simulations were run over two datasets: 
\begin{enumerate}
  \item $n=2500$ independent, standard Gaussian features $\mathbf{x}=(x_1,
    \dots, x_{50})$
  \item $n=7808$ \textit{Drosophila} ChIP-chip measurements for $23$ TFs.
\end{enumerate}
For each dataset, we drew responses $y\sim Bernoulli(\pi)$ from three generative models
corresponding to different mechanisms of action:

\begin{eqnarray}
  \pi_{AND} &=& 0.8 \cdot r_{AND}(\mathbf{x})\label{eq:and_sirf}\\
  \pi_{OR} &=& 0.8 \cdot  \left(r_{OR_0}(\mathbf{x}) + r_{OR_1}(\mathbf{x})\right)\label{eq:or_sirf}\\
  \pi _{ADD} &=& 0.4 \cdot \left(r_{ADD_0}(\mathbf{x}) + r_{ADD_1}(\mathbf{x})\right)\label{eq:add_sirf},
\end{eqnarray}
where
\begin{eqnarray}
r_{AND}(\mathbf{x}) &=& \prod_{j\in \mathcal{I}_{AND}} \mathds{1}(x_j > q_{\alpha_j})\\
r_{OR_i}(\mathbf{x}) &=& \prod_{j\in \mathcal{I}_{OR_i}} \mathds{1}(x_j > q_{\alpha_{j}})  \cdot \prod_{j\in \mathcal{I}_{OR_{1-i}}} \mathds{1}(x_j \le q_{1-\alpha_{j}}), \hspace{1em} i=0, 1\\
r_{ADD_i} &=& \prod_{j\in \mathcal{I}_{ADD_i}} \mathds{1}(x_j > q_{\alpha_j}), \hspace{1em} i=0, 1
\end{eqnarray}
and $q_{\alpha_j}$ denotes the $\alpha_j^{th}$ quantile of feature $j$. Active features $\mathcal{I}$ and quantiles were selected to ensure $\sim 10 \%$ active responses.

The three different models were designed to test a range of properties for recovered interactions. Eq. \eqref{eq:and_sirf} is an AND rule, which is active when all interacting features are highly expressed. It provides a simple baseline corresponding to a single active interacitons. Eq. \eqref{eq:or_sirf} depends on the opposing activity of two sets of features. This allowed us to evaluate the sign information recovered by siRF when the ``correct'' sign depends on the sign of other interacting features. Eq. \eqref{eq:add_sirf} corresponds to two distinct interactions that additively determine response activity. We used this model to test whether siRF can differentiate features that are interacting from features that influence responses but do not interact. 

We compared the performance of siRF, with various filtering criteria. For each dataset and generative model, we evaluated whether the methods recovered the full response generating interaction or any subset thereof. In each setting, siRF performed as well or better than iRF (Fig. S7). The improvement was greatest for models that included additive and non additive components (i.e., eq. \eqref{eq:or_sirf} and eq. \eqref{eq:add_sirf}). In these settings, filtering interactions based on null importance metrics further improved the positive predicted value of discovered interactions (Fig. S7). Scripts to run simulations and an Rmarkdown notebook with a detailed summary of simulations are available on \href{10.5281/zenodo.7992378}{Zenodo}.

\newpage
\bibliography{bibliography}

\section*{Acknowledgements}
We thank the members of the BDGP for their helpful discussions and comments. KK acknowledges the support of University of California, Berkeley, and Lawrence Berkeley National Laboratory, where he conducted work on this paper as a graduate student. BY acknowledges partial support from NSF grants 1953191 and IIS 1741340, and NSF grant 2023505 on Collaborative Research: Foundations of Data Science Institute (FODSI). SB acknowledges partial support from NSF awards DMS-1812128, DMS-2210675 and DMS-2239102; and NIH awards R01GM135926 and R21NS120227. SEC acknowledges support from (NIGMS)NIH R01-GM076655.

\section*{Author contributions statement}
All authors conceived the experiment(s). K.K. conducted the experiment(s). All authors reviewed the manuscript. 

\section*{Data and code availability}
Data and code to reproduce results in this paper are available on \href{10.5281/zenodo.7992378}{Zenodo}

\section*{Competing interests}
All authors declare no competing interests.

\newpage
\section*{Supporting figures}
\renewcommand{\thefigure}{S\arabic{figure}}
\setcounter{figure}{0}
\begin{figure}
\begin{center}
    \includegraphics[width=0.5\textwidth]{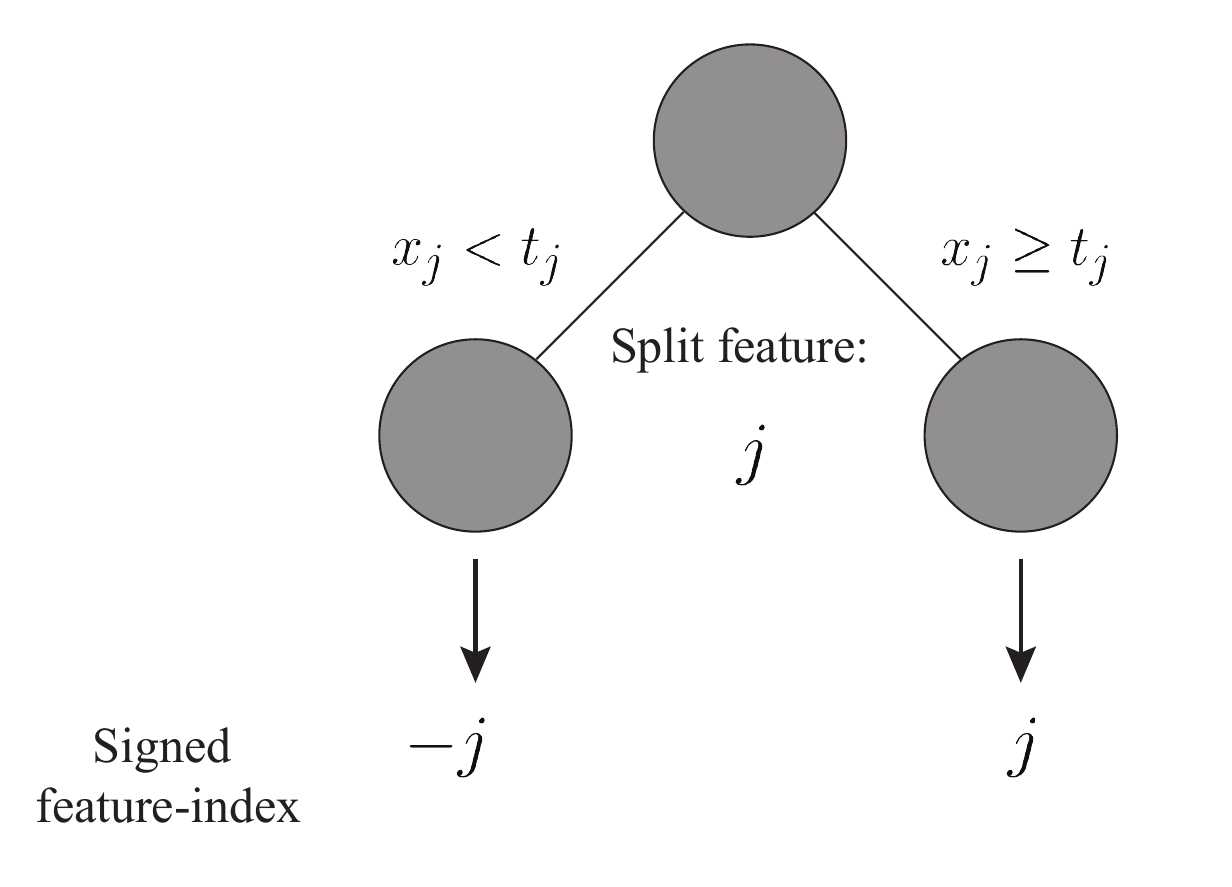}
  \caption[Signed interaction feature mapping]{Mapping a continuous feature to a signed feature index through a decision tree split. Observations that fall in the left (resp. right) node of the depicted split have feature $j$ mapped to $-j$ (resp. $+j$).
  to signed feature index.}
  \label{fig:directed}
  \end{center}
\end{figure}

\begin{figure}
  \begin{center}
    \includegraphics[width=0.95\textwidth]{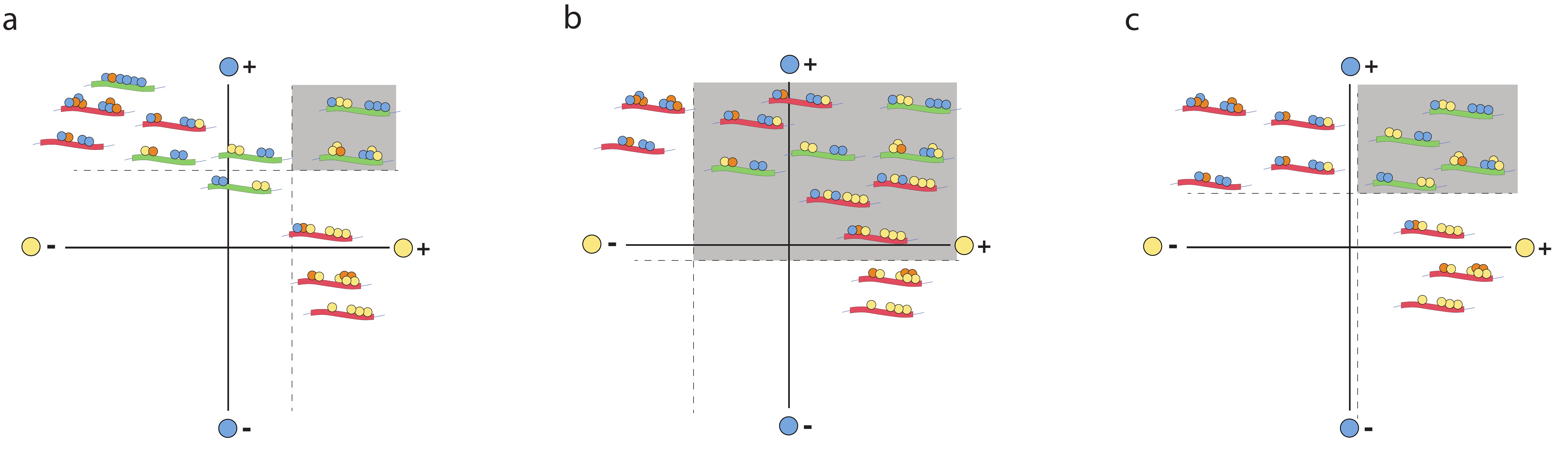}
  \end{center}
  \caption[Prevalence and precision of decision rules]{Prevalence and precision
  of decision rules. \textbf{a.} A rule with high precision but low prevalence
  captures a small portion of active responses (green segments) with high accuracy. \textbf{b.} A rule with
  high prevalence but low precision captures a large portion of active responses but also many inactive responses (red segments), resulting in lower accuracy. \textbf{c.} A rule with high precision and prevalence captures a
  large proportion of active responses with high accuracy.}
  \label{fig:prec_prev}
\end{figure}

\newpage

\begin{figure}
\begin{center}
  \includegraphics[width=0.8\textwidth]{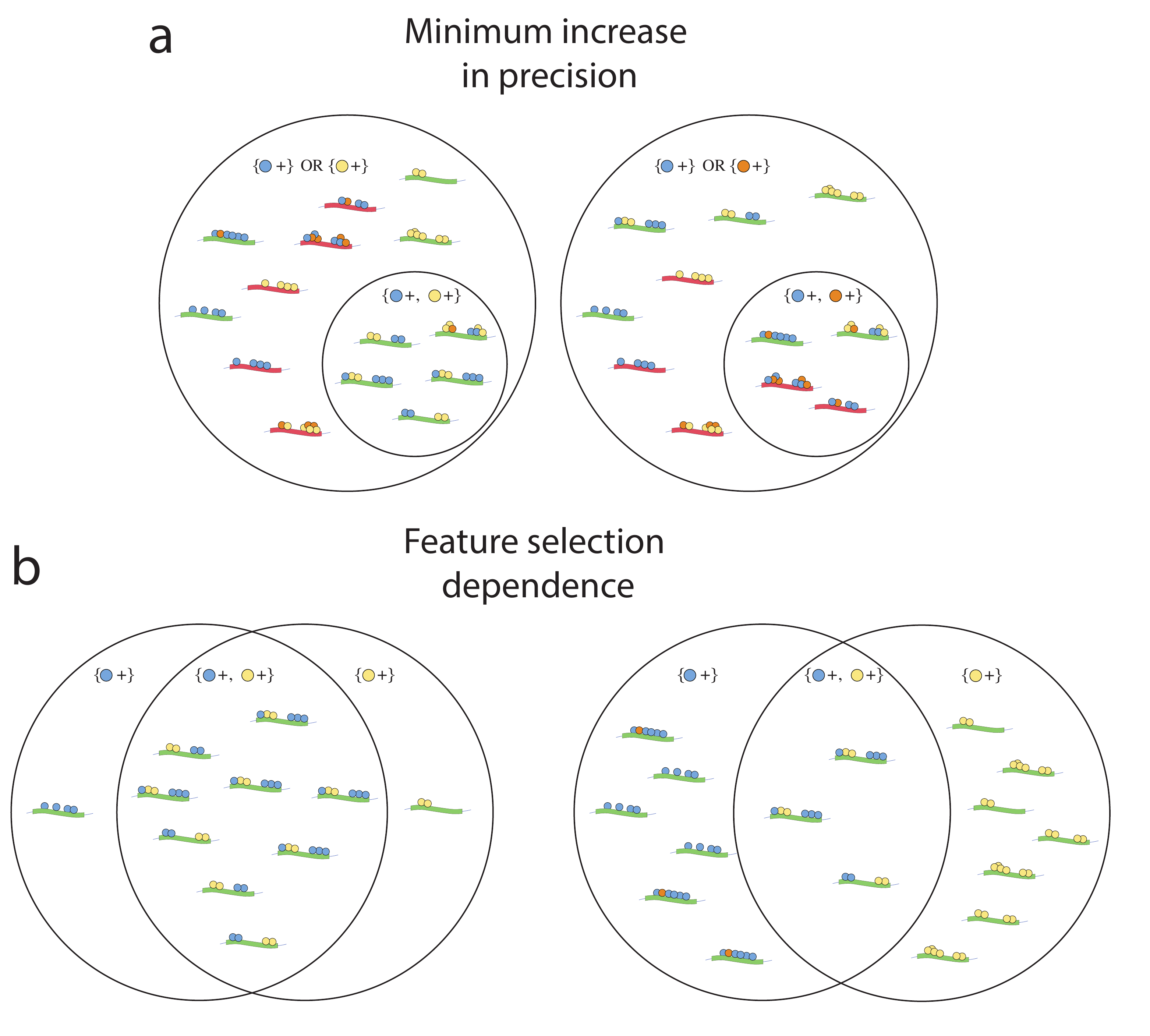}
\end{center}
  \caption[Null importance metrics]{\textbf{a.} MIP evaluates whether all features in an interaction are important for prediction. Rules characterized by high levels of both the yellow and blue TFs are more precise than those defined by high levels of the blue or yellow TFs alone. In contrast, rules characterized by high levels of both the blue and orange TFs are no more precise than those defined by high levels of the blue or orange TFs alone. \textbf{b.} FSD evaluates whether the selection of one feature in an interaction depends on the selection of the other features. On the left, high levels of the blue TF are almost always associated with high levels of the yellow TF. On the right, both the blue and yellow TFs bind many active enhancers. By chance, these TFs co-bind a small proportion of active enhancers}
\label{fig:metrics}
\end{figure}

\begin{figure}
  \begin{center}
    \includegraphics[width=0.95\textwidth]{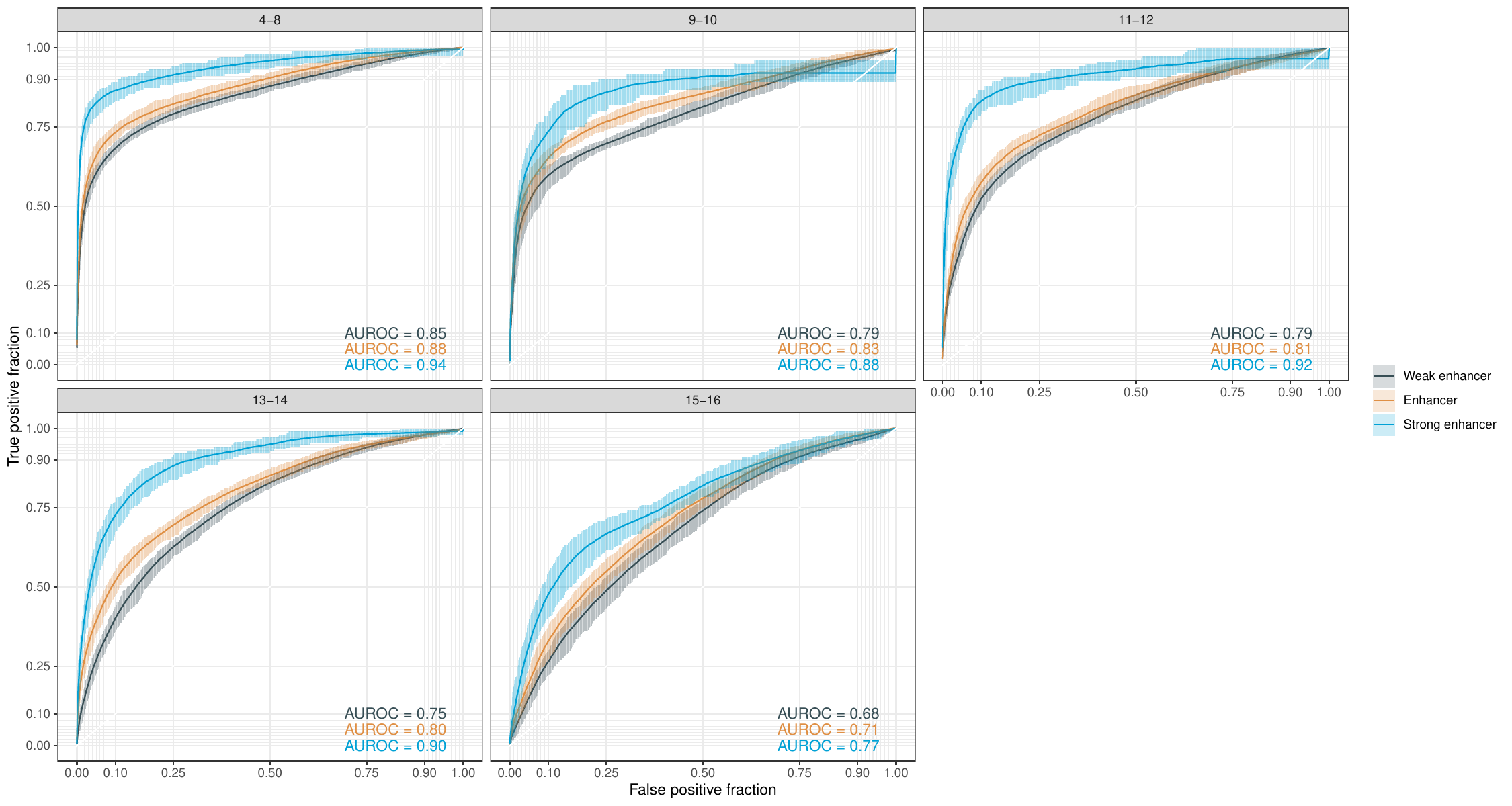}
  \end{center}
  \caption[Enhancer prediction ROC curves]{Receiver operating characteristic curves for enhancer prediction by developmental stage and enhancer activity score. Curves: average true and false positives fractions across 100 feature subsampling replicates. Error bars: $0.025$ and $0.975$ quantiles of true and false positives fractions across subsampling replicates.}
  \label{fig:roc}
\end{figure}

\begin{figure}
  \begin{center}
    \includegraphics[width=0.95\textwidth]{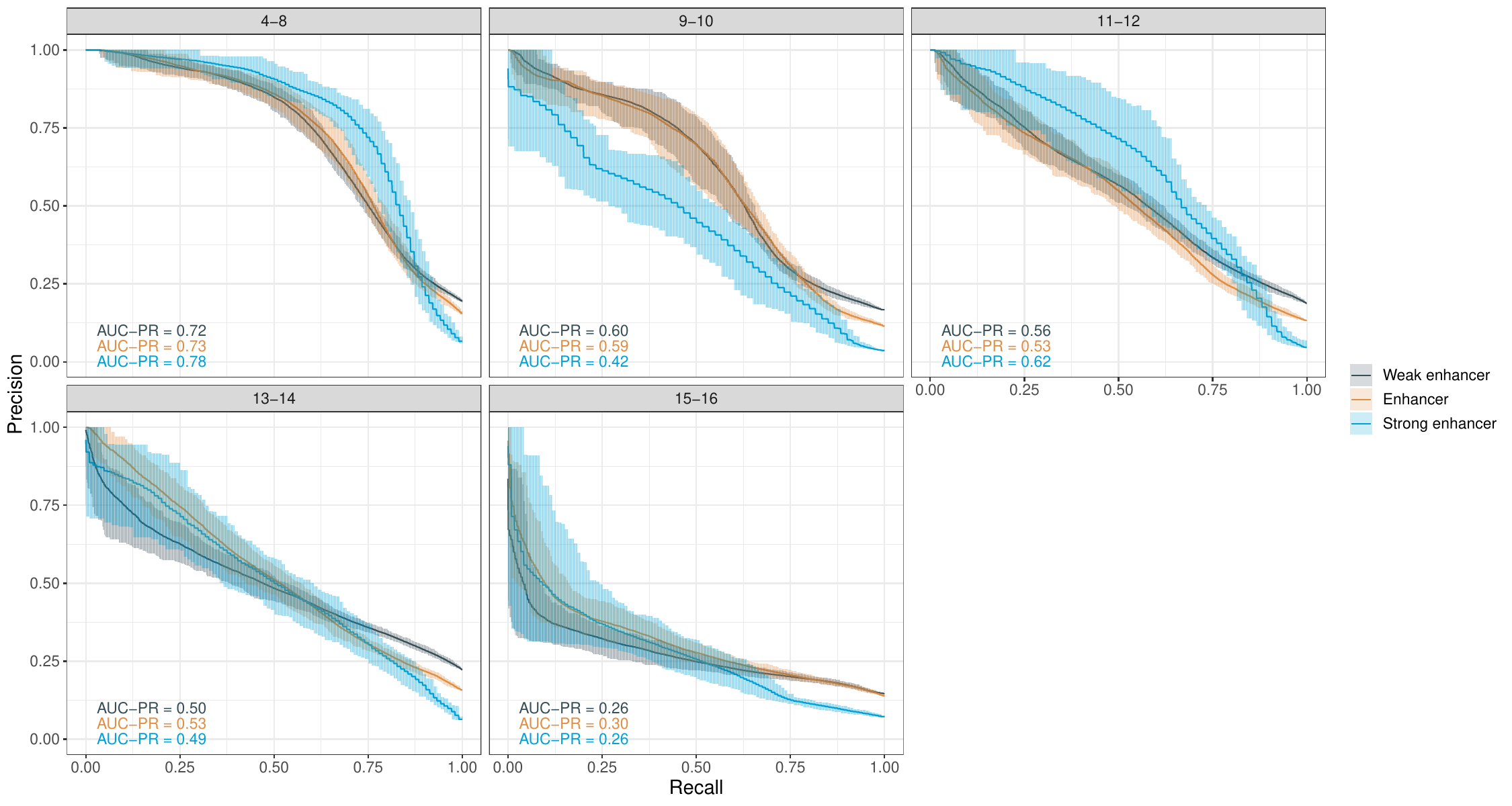}
  \end{center}
  \caption[Enhancer prediction PR curves]{Precision recall curves for enhancer prediction by developmental stage and enhancer activity score. Curves: average precision and recall across 100 feature subsampling replicates. Error bars: $0.025$ and $0.975$ quantiles of precision and recall across subsampling replicates.}
  \label{fig:roc}
\end{figure}

\begin{figure}
  \begin{center}
    \includegraphics[width=0.95\textwidth]{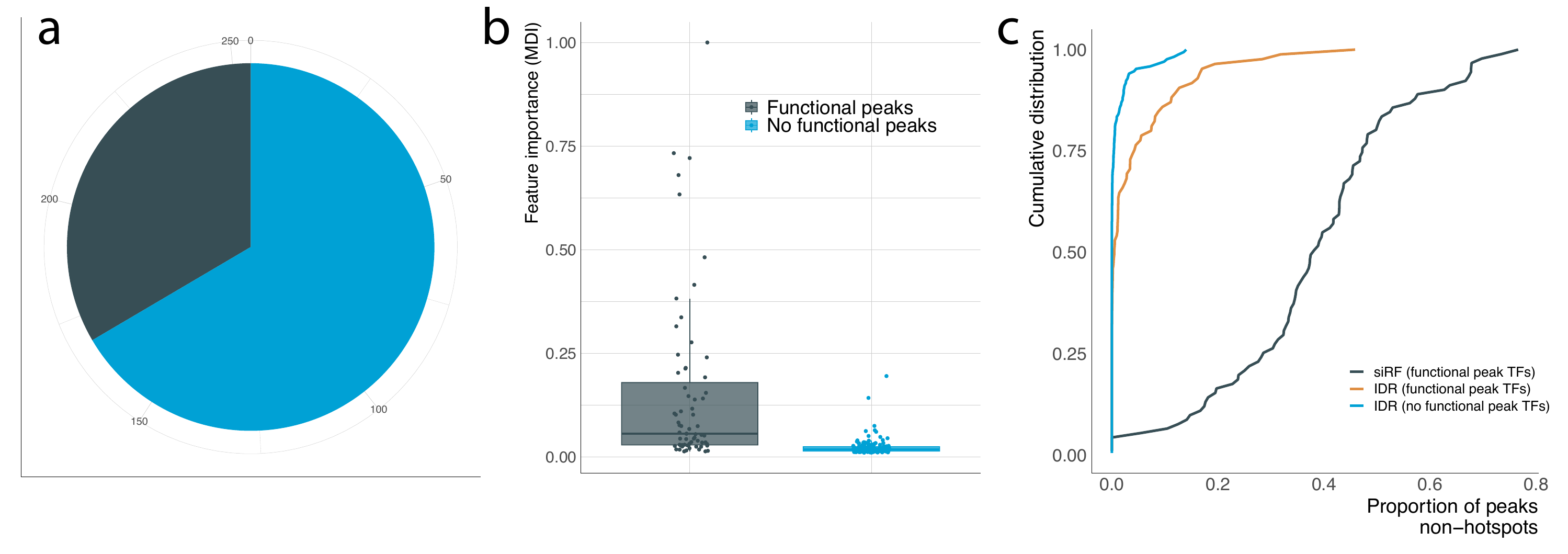}
  \end{center}
  \caption[Peak summary]{\textbf{a.} Proportion of modENCODE/modERN TFs filtered out by siRF. Color: grey = TFs with functional peaks called, blue = TFs without funcitonal peaks called. \textbf{b.} Maximum mean decrease in impurity feature importance for TFs by functional peak status. Maximum evaluated over all stages, activity thresholds, and feature subsampling replicates. \textbf{c.} Cumulative distribution in proportion of peaks localizing to non-hotspot regions. Color: grey = siRF-based functional peaks, yellow = IDR peaks for TFs with functional peaks, blue = IDR peaks for TFs without funcitonal peaks.}
  \label{fig:npeaks}
\end{figure}

\begin{figure}
  \begin{center}
    \includegraphics[width=0.95\textwidth]{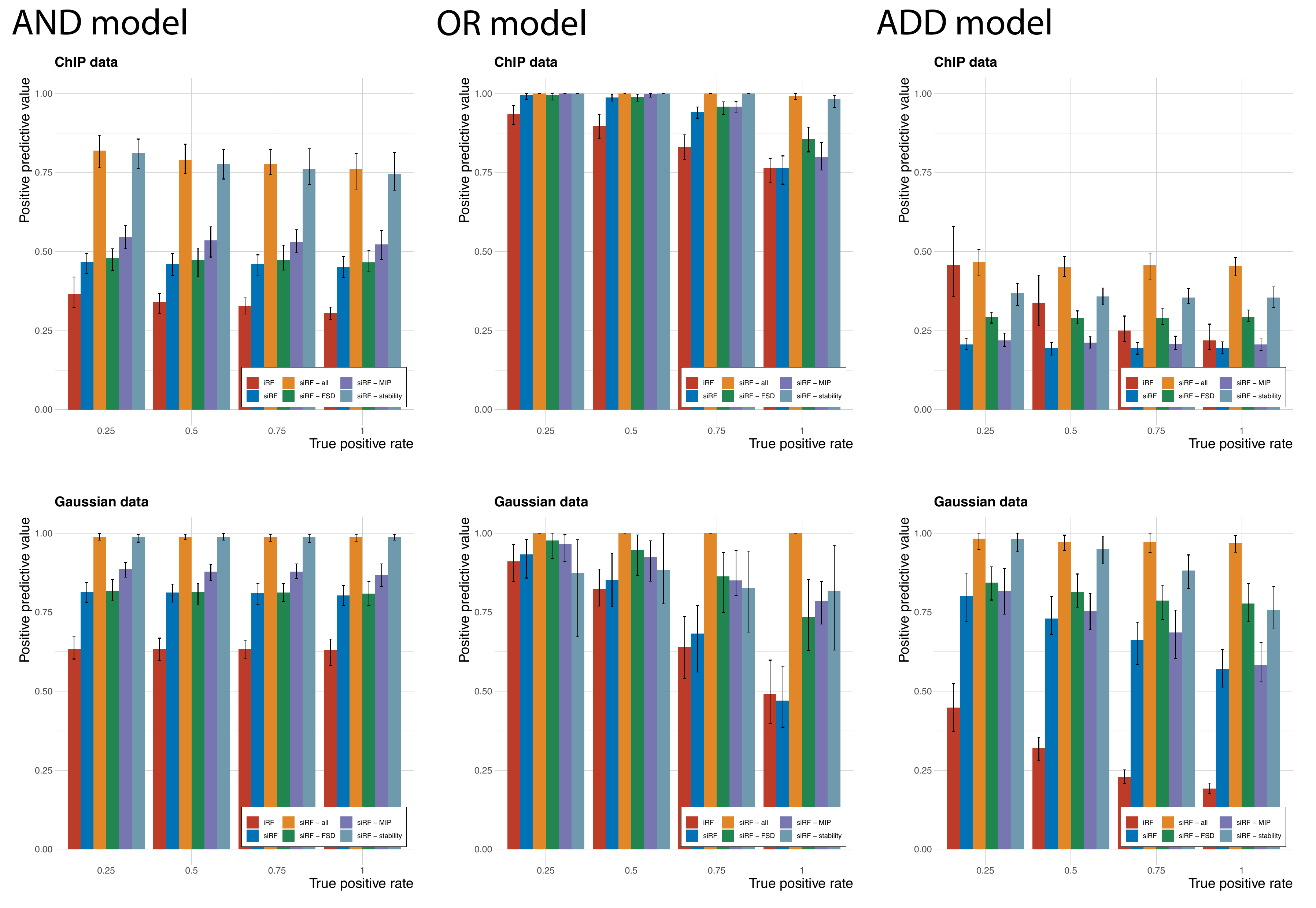}
  \end{center}
  \caption[Simulation studies]{Interaction recovery performance in simulation studies as measured by positive predictive value relative to true positive rate. Color: red = iRF, blue = siRF with no filtering, green = siRF filtering based on FSD only, purple = siRF filtering based on MIP only, grey = siRF based on stability filtering only, yellow = siRF filtering based on FSD, MIP, and stability. iRF performance evaluation does not depend on correct detection of sign.}
  \label{fig:simulations}
\end{figure}

\end{document}